\documentclass[runningheads]{llncs}
\usepackage{graphicx}
\usepackage{hyperref}
\usepackage{color}

\usepackage{graphicx}
\usepackage{latexsym}
\usepackage{bbold,amsmath,amssymb}
\usepackage{amsmath,amssymb,bbm}

\newcommand{\R}{\mathbb{R}}
\newcommand{\E}{\mathbb{E}}
\newcommand{\Ucal}{\mathcal{U}}

\newcommand{\Fcal}{\mathcal{F}}
\newcommand{\truncexp}{\mathcal{E}}

\newcommand{\one}{\mathbb{1}}
\begin{document}

\title{Optimal time sampling in physics-informed neural networks}
\titlerunning{Optimal time sampling PINN}
\author{Gabriel Turinici\orcidID{0000-0003-2713-006X}}
\authorrunning{G. Turinici}
	\institute{CEREMADE, Universit\'e Paris Dauphine - PSL,\\ Place du Marechal de Lattre de Tassigny,\\ Paris 75016, FRANCE \\ 
		\email{Gabriel.Turinici@dauphine.fr}\\
		\url{https://turinici.com}}
	\maketitle              
\begin{abstract}
Physics-informed neural networks (PINN) is a extremely powerful paradigm used to solve equations encountered in scientific computing applications. An important part of the procedure is the minimization of the equation residual which includes, when the equation is time-dependent, a time sampling. It was argued in the literature that the sampling need not be uniform but should overweight initial time instants, but no rigorous explanation was provided for this choice. In the present work we take some prototypical examples and, under standard hypothesis concerning the neural network convergence, we show that the optimal time sampling follows a (truncated) exponential distribution. In particular we explain when is best to use uniform time sampling and when one should not. The findings are illustrated with numerical examples on linear equation, Burgers' equation and the Lorenz system. 
\end{abstract}

\section{Introduction and literature review}

Following their recent introduction in~\cite{RAISSI2019686},  
physics-informed neural networks became a powerful tool invoked in scientific computing 
to numerically solve ordinary (ODE) or partial  (PDE) differential equations in physics \cite{penwarden_unified_2023} 
including  high dimensional (e.g. Schrodinger) equations~\cite{hu2024tackling}, finance \cite{bae_option_2024_pinn,pinn_vol_surface24}, control problems~\cite{mfg_pinn24}, data assimilation and so on. As such it became an important framework that leverages the power of neural networks (NN). Even if successful applications are reported for many situations encountered in numerical simulations, however the workings of PINNs are not yet fully optimized and research efforts are nowadays targeted towards improving the output quality or training process, cf. \cite{pinn_siam21} and related works.

We will focus here on time-depending equations that can be formalized as solving
\begin{eqnarray}
& \ &  
\partial_t u(t,x) = \Fcal (u) , \label{eq:general_equation}
\\ & \ & u(0,x) = u_0, \ \forall x \in \Omega 
\\ & \ & u(t,x) = u_b(t,x), \ \forall x \in \partial \Omega,  
\forall t \in [0,T], 
    \label{eq:general_equation3}
\end{eqnarray}
where $\Fcal$ is an evolution operator (see below for examples), $u$ is the unknown, $u_0$ the initial condition, $\Omega$ the spatial domain and 
$u_b(t,x)$ the conditions at the boundary $\partial \Omega$ of $\Omega$;
this can be supplemented with additional measured quantities or physical information in data assimilation settings (not used here). We will not investigate in this work the difference between weak and strong formulations so we suppose  \eqref{eq:general_equation}  is true 
pointwise with classical time and partial derivatives (classical solutions). Note that when
$\Omega$ is a discrete set \eqref{eq:general_equation}  becomes a ordinary differential equation (ODE) and in this case $\partial_t$ is to be replaced by $d/dt$.

\begin{figure}
\centerline{\includegraphics[width=0.5\textwidth]{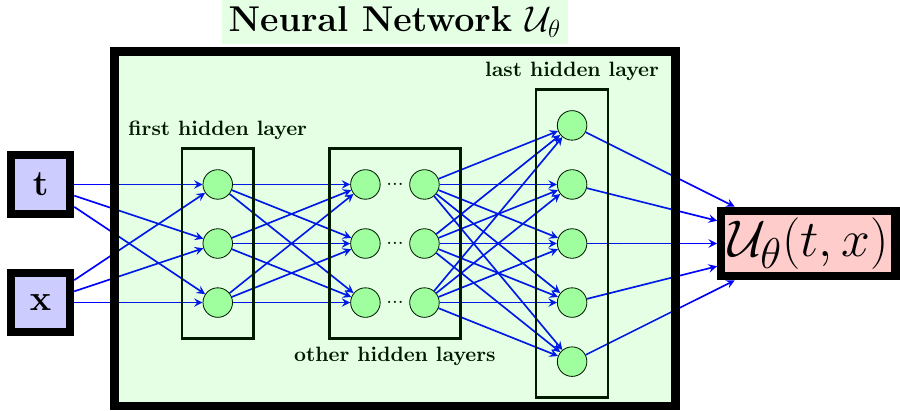}}
\caption{An illustration of network $\Ucal_\theta$. It takes as input a time $t$ and a space value $x$ and outputs the solution candidate  $\Ucal_\theta(t,x)$ for this input couple. The NN is trained so that $\Ucal_\theta(t,x)$ is close to the solution $u$ of \eqref{eq:general_equation}.} \label{fig:pinn_nn}
\end{figure}

In its simplest form, the PINN approach constructs a  neural network $\Ucal$ indexed by parameters $\theta$ mapping the input $(t,x) \in [0,T] \times \Omega$ to 
$\Ucal_\theta(t,x) \in \R$ that will stand for the (unknown) solution $u(t,x)$ (see figure \ref{fig:pinn_nn} for an illustration). To ensure that $\Ucal_\theta$ is close to $u$ the following functional is minimized with respect to 
$\theta$ by usual means of deterministic or stochastic optimization as is classically done for NNs~:
\begin{eqnarray}
& \ & 
    L(\theta) := \int_0^T \int_\Omega E_\theta(t,x)^2 dx dt + 
    \label{eq:definition_loss_error_term}
\\ & \ & 
    c_{ic} \int_\Omega (\Ucal_\theta(t,x) - u_0(x))^2dx +
    c_{bc}  \int_0^T \int_{\partial \Omega} (\Ucal_\theta(t,x) - u_{bc}(t,x))^2dx dt.
\end{eqnarray}
Here $c_{ic}$, $c_{bc}$ are some positive coefficients and $E_\theta(t,x)$ is the error term~:
\begin{equation}
    E_\theta(t,x) = \partial_t \Ucal_\theta(t,x) - \Fcal (\Ucal_\theta)(t,x).
\end{equation}

Assuming that the NN is expressive enough i.e., that the true solution $u$ belongs to the set of all possible NN mappings $\Ucal_\theta$ then the minimizer  of the loss is exactly the solution $u$.
The  integral terms of $L(\theta)$ are generally computed through either collocation points or random sampling. 
The focus of this paper is on the computation of the terms involving the time integral and more precisely we will mostly investigate the term in \eqref{eq:definition_loss_error_term}.

The choice of the collocation points has an impact on the  efficiency of  the PINN result. For instance in  \cite{subramanian2022adaptivePINN} the authors argue that 
adapting the location of these points by over-weighting
areas where $E_\theta$ is large will improve the outcome. 
In another approach, \cite{wight2020solving} 
C.L. Wight and J. Zhao propose several adaptive sampling in space and time; among their proposals is the time marching where the time interval 
$[0,T]$ is divided into  segments solved sequentially (their ``approach II"); a different approach considers a total span 
$[0,t]$ that is increased from a small value $t$ to the target value $T$ (``approach I''). 
 Asking the same question, M. Penwarden and co-workers proposed in a very recent work \cite{penwarden_unified_2023}
``a stacked-decomposition method" that combines time marching with a form of 
Schwartz overlapping time domain decomposition method.
 Investigation of time sampling also lead S. Wang, S. Sankaran and P. Perdikaris
 \cite{pinn_causality} to introduce the causality concept where it is recognized that error made earlier in the time interval will escalate to the final time $T$; they propose to over-sample points close to $0$ and decrease the sampling weight as time progresses. 
 Our contribution lies very much within this line or thought but here 
 we give rigorous insights into several points~:
 \begin{itemize}
 \item what is the best functional form for the decrease in sampling weights from $0$ to $T$  
 \item for what problems is this causal sampling likely to give best results and where is it less critical ? In particular are there any situations where  it is optimal to under-weight points near $0$ and over-weight points near $T$ ? 
 \item how is this related to the optimization procedure used to minimize the loss functional.
 \end{itemize}
The balance of the paper is as follows : in section 
\ref{sec:notations_and_theory} we introduce notations and give theoretical insights that we illustrate in section~\ref{sec:numerical}  with numerical results; we conclude with remarks in section \ref{sec:discussion}.

\section{Theoretical setting and results} \label{sec:notations_and_theory}
To explain the sampling / weighting question, we consider here the simplest possible setting, that of a linear ODE~: $\text{cardinality}(\Omega)=1$, $\Fcal(u) = \lambda u$~:
\begin{equation}
    u'(t) = \lambda u(t), \  u(0) = u_0.
    \label{eq:odelambdadef}
\end{equation}
The network $\Ucal$ parameterized by some $\theta$ maps any input $t\in [0,T]$ into the value $\Ucal_\theta(t)$. We want the mapping $t \mapsto \Ucal_\theta(t)$ to represent the solution to 
\eqref{eq:odelambdadef} and in this case 
the equation error is
\begin{equation}
    f_\theta(t) := \Ucal'_\theta(t) - \lambda \Ucal_\theta(t).
    \label{eq:def_ftloss}
\end{equation}

To simplify again our setting we will not describe the treatment of the initial condition $u_0$ and instead assume the network outputs some function that  already has $\Ucal_\theta(0)=u_0$. Note that in this simple setting one can easily ensure this equality by just shifting the output 
\begin{equation}
\Ucal_\theta(t) \mapsto \Ucal_\theta(t) - \Ucal_\theta(0) +u_0.    
\label{eq:trick_initial_cond}
\end{equation}
 Similar techniques have been used in the literature, see \cite[section E]{pinn_causality}. 
Then the PINN method prescribes to minimize, with respect to $\theta$, the following loss function~:
\begin{equation}
    L_\rho(\theta) := 
    \E_{t \sim \mu} f_\theta(t)^2 = 
    \int_0^T f_\theta^2(t) \rho(t) dt.
    \label{eq:loss_rho}
\end{equation}
The loss is the second moment of the equation error $f_\theta$
introduced in {\eqref{eq:def_ftloss}. Here
 time $t$ follows a probability law $\mu$  
supported in $[0,T]$ and 
density $\rho(\cdot)$; we write $\mu(dt) = \rho(t) dt$. 

The final goal is to obtain a good approximation of the final solution 
$u(T)$ which is the  unknown and the main goal of the procedure. So the real quantity to be minimized is
$|u(T)-\Ucal_\theta(T)|$ but this cannot be done directly because $u(T)$ is not known.

\subsection{Model for computational resources}

To find the solution one uses  (stochastic or deterministic) optimization algorithms, most of them derived from  the initial proposal of Robbins and Monro \cite{robbins_stochastic_1951}) that was called latter Stochastic Gradient Descent. In turn this was followed by a large set of proposals used nowadays in neural network optimization (Nesterov, momentum, Adam, RMSprop, etc). The deterministic counterpart algorithms  (gradient descent, BFGS, L-BFGS and so on) appear on the other hand in standard textbooks \cite{numerical_recipes_2007}.
These algorithms find the solution in an iterative manner and convergence is ensured only in the limit of infinite iterations. So  we never have the exact solution but some approximation of it. Moreover the computational resources are not infinite either so in practice one is limited by the available resources (in wall clock time or in total operations count or in any other metric). 
In particular, smaller is the absolute value of $f_\theta(\cdot)$ more computational resources are consumed.

To model this cost we refer to general results on the convergence of optimization algorithms. 
The convergence of the stochastic and deterministic procedures has been analyzed  in detail see~\cite{chen_stochastic_2002} for a classic textbook and \cite{mertikopoulos_almost_2020,sgd_conv_non_cx20} for recent works or self-contained proofs  \cite{gabriel_turinici_convergence_2023,anita2024convergence}. It was proved that, in general, the convergence to the exact solution occurs at various speeds including quadratic or exponential convergence. 
The most often the square 
of the error is of order $O(\frac{1}{n})$ where $n$ is the number of iterations, proportional to the numerical effort.  
For convenience we will denote from now on the error by $w(t)$ so finally we have that the square of the error is, say, of order 
$w(t)^2 \sim O(\frac{1}{n})$. 
So, if we take as a constraint that the total numerical cost is bounded by some $B\ge 0$, 
the error $w(t)$ will be associated to a cost of order $1/w(t)^2$ so
the optimization algorithm will find some error 
$w(t)$
that satisfies 
\begin{equation}
    \int_0^T \frac{1}{w(t)^2} dt \le B. 
    \label{eq:error_definition}
\end{equation}
Of course, the exact functional form of \eqref{eq:error_definition} is subject to discussion and e.g., when exponential convergence occurs we will rather have 
\begin{equation}
    \int_0^T -\ln(w(t)^2) dt \le B. 
    \label{eq:error_definition_exp}
\end{equation}
In fact the arguments below apply to both such formulations and to many other also, so for simplicity we will suppose \eqref{eq:error_definition} is true. 

To conclude, if computational resources are limited by a total amount $B$ we only have access to  errors $w(t)$ that satisfy 
\eqref{eq:error_definition} and not better. The question is how to choose $w$ to minimize the final error between the numerical and the exact solution and how does the time sampling enters into this quest. 

Some remarks remains still to be made at this juncture: when minimizing some loss functional under computational constraints \eqref{eq:error_definition}, it may happen that several values give the same loss level and same computational cost; in this case we cannot be sure which one we will get. So, in a prudent stance, we will suppose from now on that
\begin{quote} 
{\bf (H Opt)}~: The computational procedure results in some error level $w(\cdot)$ that minimizes the loss functional under constraint \eqref{eq:error_definition}. 
 If several errors give the same cost and loss level we will assume the worse one is actually obtained.
\end{quote}

\subsection{Step 1: overall optimality}
We give here a first result that will be an lower bound on the error $|\Ucal_\theta(T)- u(T)|$.

\begin{proposition}
Denote 
\begin{equation}
w(t) := \Ucal'_\theta(t) - \lambda \Ucal_\theta(t),
\label{eq:definition_w}
\end{equation}
 and assume that \eqref{eq:error_definition} holds true. Then under hypothesis {\bf (H Opt)}
 the error $|\Ucal_\theta(T)- u(T)|$ is at least equal to
\begin{equation}
\frac{1}{B^{1/2}} \left(\frac{3(e^{2\lambda T/3}-1)}{2 \lambda} \right)^{3/2},
\label{eq:minimum_error_B}
\end{equation}
with equality when 
 $w(t)$ is proportional to $e^{-\lambda(T-t)/3}$.
\label{prop:minimal_error}
\end{proposition}
\noindent {\bf Proof~:} 
We deal here with an optimization problem and need to find the minimum value of the error under resources constraints \eqref{eq:error_definition}.
In general this can be formulated as a Euler-Lagrange constraint optimization problem. But in this particular case it can be settled more directly. 
Let us  first write the definition
\eqref{eq:definition_w}
of $w$ as~:
$	\Ucal'_\theta(t) = \lambda \Ucal_\theta(t) + w(t)$.
Then, denoting $\delta u(t) = \Ucal_\theta(t) - u(t)$ we can write 
$\delta u (t)' = \lambda \delta u (t) + w(t)$,
or, by using classical formulas for the solution of this equation~:
\begin{equation}
	|\Ucal_\theta(T)- u(T)| = 
	|\delta u (T)| =  \left|\int_0^T e^{\lambda(T-t)}w(t)dt\right|.
	\label{eq:integral_form_of_errorT}
\end{equation}
Of course, the worse case is realized when $w(s)$ is positive and then we will have 
$|\Ucal_\theta(T)- u(T)| =\int_0^T e^{\lambda(T-t)}w(t)dt$.
Use now the H\"older inequality
for the functions
$ (e^{\lambda(T-t)}w(s))^{2/3}$ and
$w(s)^{-2/3}$
and exponents $p=3/2$, $q=3$~:
\begin{eqnarray}
	& \ & 
	\int_0^T e^{2\lambda(T-t)/3}dt =
	\int_0^T (e^{\lambda(T-t)}w(t))^{2/3} \cdot  w(t)^{-2/3}dt
	\nonumber \\& \ & 
	\le 
	\left(
	\int_0^T e^{\lambda(T-t)}w(t)dt \right)^{2/3}
	\left(\int_0^T \frac{1}{w^2(t)}dt \right)^{1/3}
	\le
	B^{1/3} \left(\int_0^T e^{\lambda(T-t)}w(t)dt 
	\right)^{2/3}.
\end{eqnarray}
It follows that 
$ \int_0^T e^{\lambda(T-t)}w(t)dt 
\ge \frac{1}{B^{1/2}} \left(\frac{3(e^{2\lambda T/3}-1)}{2 \lambda} \right)^{3/2}$.
Equality occurs when 
$e^{\lambda(T-t)}w(t)$ is proportional to 
$\frac{1}{w^2(t)}$ which means that $w(t)$ is proportional to $e^{-\lambda(T-t)/3}$.

\begin{remark}
The proof technique here works also for time-depending $\lambda$.
\end{remark}

\subsection{Optimal time sampling distribution}
The proposition 
\ref{prop:minimal_error}
states that, at resources level $B$ one cannot do better than
\eqref{eq:minimum_error_B}. The question is how can one choose the right $\rho$ in order to reach this minimal error level. The answer is in the next result. 

\begin{proposition}
Under the hypothesis {\bf (H Opt)} the minimization problem corresponding to the loss $L_\rho$ in \eqref{eq:loss_rho} and the computational constraint
\eqref{eq:error_definition}
is guaranteed to obtain the best error level of  proposition \ref{prop:minimal_error} only when 
 $\rho$ is the density of the exponential truncated law
$\truncexp^{0,T,4 \lambda/3}$.
\label{prop:optimality}
\end{proposition}
\noindent {\bf Proof~:}
Let us write 
\begin{eqnarray}
	& \ & 
	\left(
	\int_0^T \rho(t)^{1/2}\right)
	\le \left(\int_0^T \frac{1}{w^2(t)}dt \right)^2
	\left( \int_0^T w^2(t) \rho(t)dt \right)^2
	\nonumber \\& \ & 
	\le 
	\left(\int_0^T \frac{1}{w^2(t)}dt \right)^2
	\left( \int_0^T w^2(t) \rho(t)dt \right)^2
	\le 
	B^2 \left(\int_0^T w^2(t) \rho(t)dt \right)^2 = B^2 L_\rho^2.
	\nonumber \\& \ & 
\end{eqnarray}
The loss will be minimized when there is equality in the above inequality which means that 
$1/w^2$ is proportional to $w^2 \rho$ that is $\rho$ is proportional to $w^{-4}$. 
On the other hand, given proposition \ref{prop:minimal_error}, 
the proof of proposition \ref{prop:optimality}
is a matter of asking for which $\rho$  the minimizer $w(s)$  of 
$L_\rho$ under constraint \eqref{eq:error_definition}
will 
be proportional to $e^{-\lambda(T-t)/3}$.
Putting together these two arguments we obtain that overall error loss $|\Ucal_\theta(T) - u(T)|$ is minimized when $\rho$ is proportional to $e^{4\lambda(T-t)/3}$
i.e., it corresponds to the truncated law  
$\truncexp^{0,T,4 \lambda/3}$.

\begin{remark}
So we proved that under hypothesis {\bf (H Opt)} concerning the algorithm's convergence speed the error is minimal when the time sampling follows a truncated exponential law. The same result holds true if instead we consider algorithms with exponential convergence \eqref{eq:error_definition_exp}, see remark in the beginning of the proof of proposition~\ref{prop:minimal_error}.
\end{remark}

\subsection{Remarks on general settings: regimes of Lyapunov exponents}

The example \eqref{eq:odelambdadef} may seem somehow too simple but in fact covers many situations encountered in practice. To this end let us recall the concept of {\it maximal Lyapunov exponent}  used in the study of dynamical systems, particularly in chaos theory, to characterize the behavior of trajectories. The maximal Lyapunov exponent quantifies the rate of exponential divergence or convergence of nearby trajectories in the system.

Consider a dynamical system described by ordinary differential equation
 $   u'(t) = \mathfrak{F}(t,u(t))$.
If one considers two similar initial conditions $u(0)$  and $u(0)+\delta u(0)$ with $\delta u(0)$ playing the role of a small perturbation, the distance between these trajectories evolves over time according to the linearized dynamics around the system's trajectory. Specifically, if $\delta u(t)$ represents the perturbation at time $t$, then~:
$\delta u(t)' = \nabla_u \mathfrak{F}(t,u(t)) \delta u(t)$. 
The maximal Lyapunov exponent $\lambda$ (see \cite[section 5.3]{chaos_textbook}) is defined (for dimension $1$) as the average exponential rate of separation of those trajectories:
$\lambda =\lim_{t\to \infty } \lim_{\|\delta u(0)\|\to 0} \frac{1}{t} \ln \left(\frac {\|\delta u(t)\|}{\|\delta u(0)\|} \right)$.  
This can be also read as 
$\|\delta u(t)\| \sim e^{\lambda t} \|\delta u(0)\|$.
The maximal Lyapunov exponent characterizes the system's sensitivity to initial conditions and can provide insights into whether the system exhibits chaotic behavior. In particular~:
\begin{enumerate}
\item if $\lambda >0$
close trajectories diverge exponentially, indicating chaotic behavior; in this case truncated exponential time sampling (weighting) is mandatory as it will be seen in numerical examples~;
\item  if $\lambda<0$ close trajectories converge exponentially, suggesting stability. In this case no special time sampling or weighting seems necessary and one may even imagine that an inverse sampling that puts more weight on larger time values can do better because initial perturbations fade away exponentially fast. This would correspond to parabolic evolution like the heat equation. In this situation the system evolves towards a static equilibrium.

\item  if $\lambda=0$  trajectories neither converge nor diverge, indicating a marginally stable or periodic behavior. Here for safety some truncated exponential sampling can be enforced.
Systems involving  conservation laws (not evolving towards static equilibrium) are present in this case (for PDE this will be called hyperbolic equations). 
\end{enumerate}
Note that if one is interested only in what happens in a neighborhood of the initial point and for small times, the corresponding concept is the {\it local Lyapunov exponent}  which is related to the spectrum of the Jacobian $\nabla \mathfrak{F}$; note that this local metric can change regime as happens for instance with the Lorentz system.

\section{Numerical examples} \label{sec:numerical}

All these numerical tests are reproducible using codes available on Github~\footnote{\scriptsize\url{https://github.com/gabriel-turinici/pinn_exponential_sampling} version August 31st 2024.}.

\subsection{Example in section~\ref{sec:notations_and_theory}}
\label{sec:numerical_lambda}
We investigate first the example in section~\ref{sec:notations_and_theory} and describe below the numerical parameters and setting.

\subsubsection{NN architecture and training parameters}
A neural network is used that will construct the mapping from $t,x$ to the numerical solution $\mathcal{U}_\theta(t)$ as in figure~\ref{fig:pinn_nn}. When the spatial dimension is not present, as is the case in section~\ref{sec:notations_and_theory}, then $\mathcal{U}_\theta(t)$ has a single input which is the time $t$. The NN has 5 fully connected (FC) layers with Glorot uniform initialization seeded with some constant for reproducibility; we checked that any other seed gives similar results. Each layer has $10$ neurons  and 'tanh' activation. This activation is classical in PINN because ReLU would give null second order derivatives. Then a final FC layer with no activation and one final neuron is used to output the model prediction.

We set $T=1$. 
The initial value $u_0$ is taken to some non-special value i.e. not $0$ or $1$; here $u_0=\sqrt{15}$ but many other values have been tested and give similar results. The shifting trick 
\eqref{eq:trick_initial_cond}
in section~\ref{sec:notations_and_theory} is used to be sure that the initial condition is not an issue and will be respected exactly (otherwise one has to study also the impact of the regularization coefficient used to impose the initial condition). The loss function \eqref{eq:loss_rho} is employed and the sampling is performed with a truncated exponential law of rate $r$ which is not necessarily equal to $\lambda$ (recall that in general $\lambda$ is unknown). The loss is computed by taking $100$ sampling points which are exactly the quantiles of the law 
$\truncexp^{0,T,r}$, see formula \eqref{eq:quantiles_truncated_exp}. We take $N_{iter}=500$ iterations and an Adam optimizer with default learning rate (in our TensorFlow 2.15.0 version this is $10^{-3}$).
We checked that the quality of the numerical results can be made better by taking more iterations.

\subsubsection{Validation of the NN architecture : expressiveness}
\begin{figure}
\centerline{\includegraphics[width=0.75\textwidth]{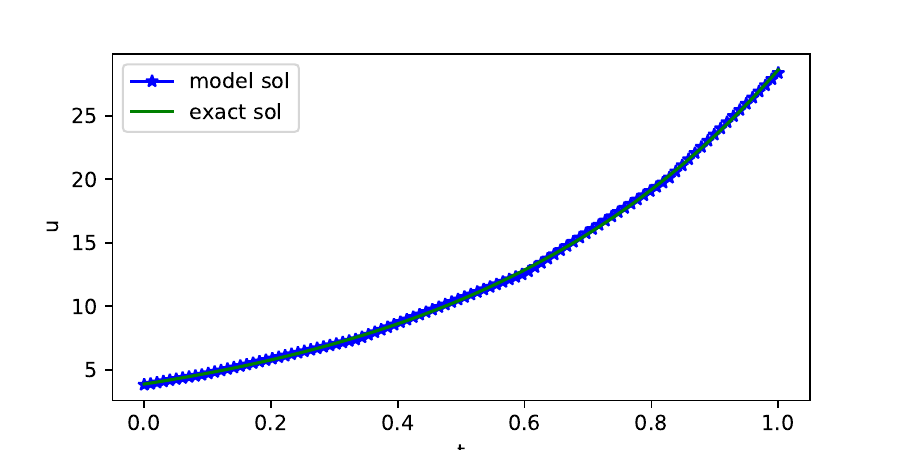}}
\caption{Test of the model expressiveness (results for $\lambda=2.$). The model solution is graphically indistinguishable from the exact solution meaning that the NN is complex enough to reproduce the shape of the solution.} \label{fig:lambda2_expressivity}
\end{figure}
We checked first that the model is expressive enough. This means that, without any PINN framework, we just checked that the NN architecture can produce functions close enough to the exact solution $u(t)$. We used Adam optimizer with default parameters and mean square error between the model and the known exact solution. The results are plotted in figure~\ref{fig:lambda2_expressivity} and show that the model is indeed expressive enough. Of course, this is just a theoretical possibility as the exact solution is in practice unknown and has to be found through the minimization of the PINN loss functional. But it still says that a good design of the loss functional should give good numerical results, i.e. that the model architecture is not the limiting hyper-parameter.

\subsubsection{Validation of the PINN procedure: solution quality}

We now run the main PINN code. The solutions obtained 
for $\lambda=2.0$
are plotted in figure \ref{fig:lambda2_sol}. It is seen that the model is giving a good solution. This solution appears not enough converged for 500 epochs so we also gave the result for 1500 epochs where the numerical and exact solutions are indistinguishable graphically. This means that the PINN procedure is sound and gives expected results, in line with the literature.
\begin{figure}
\centerline{
\includegraphics[width=0.49\textwidth]{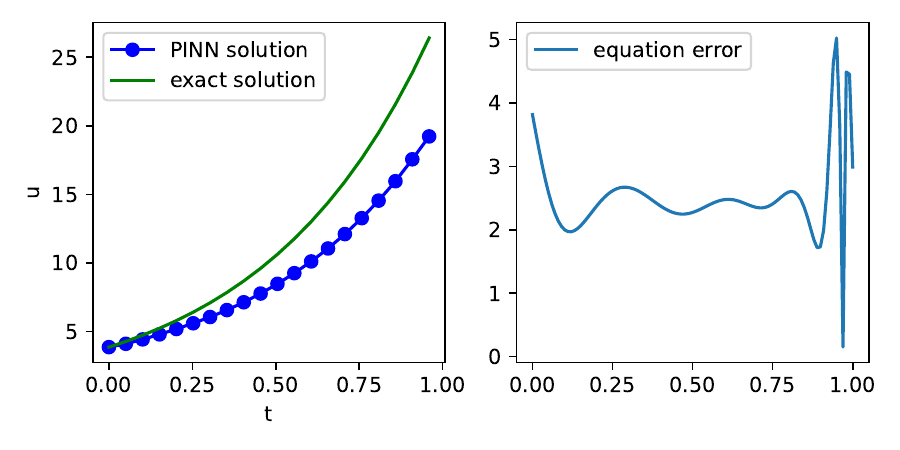}
\includegraphics[width=0.49\textwidth]{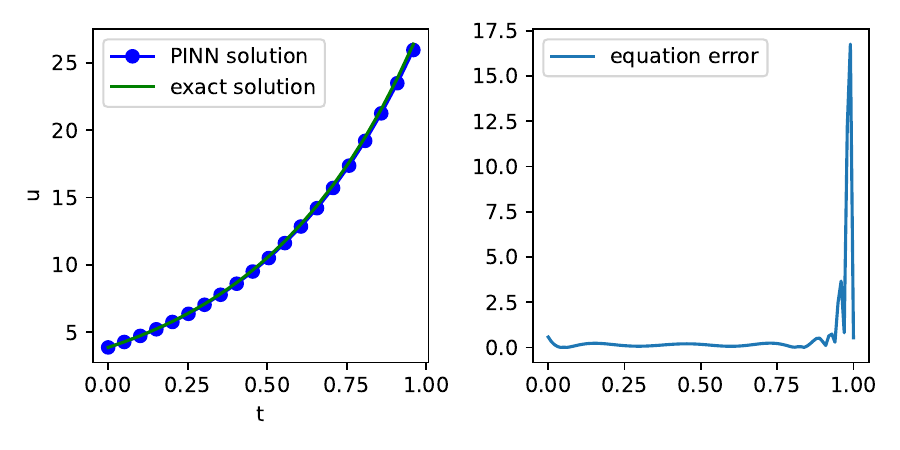}}
\caption{Results for $\lambda=2.$ and sampling parameter $r=2.0$. First two plots: the results for $500$ epochs. Last two plots: results for $1500$ epochs.} \label{fig:lambda2_sol}
\end{figure}

\subsubsection{Numerical results: solution sampling influence}

\begin{figure}[htpb!]
\hbox{\hspace{2.7cm} $\lambda=2$ \hspace{5.2cm}  $\lambda=-2$}
\centerline{
\includegraphics[width=0.49\textwidth]{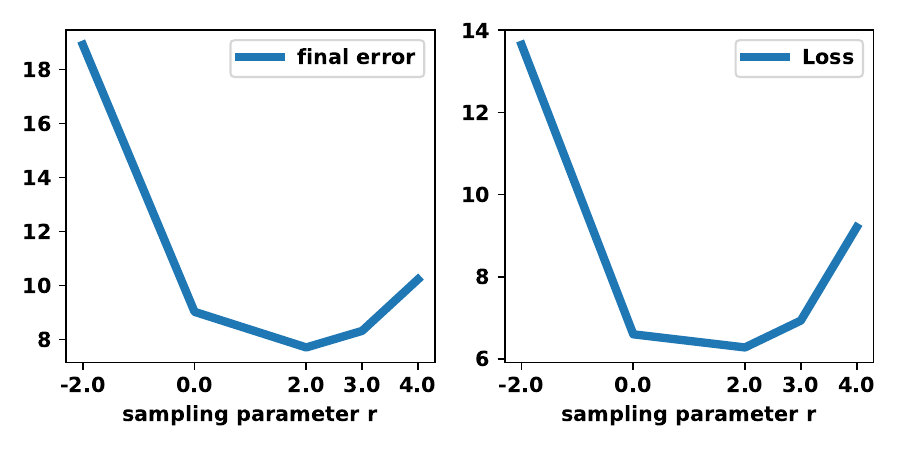}
\includegraphics[width=0.49\textwidth]{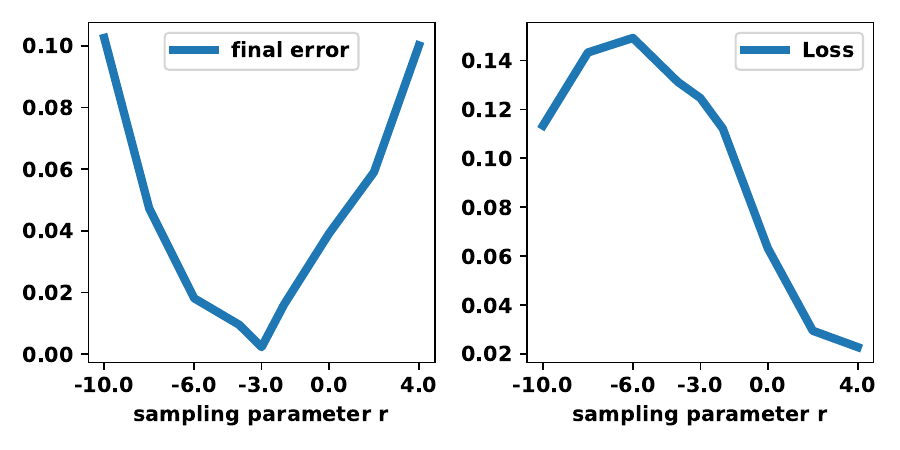}}

\hbox{\hspace{5.7cm} $\lambda=0$}
\centerline{
\includegraphics[width=0.5\textwidth]{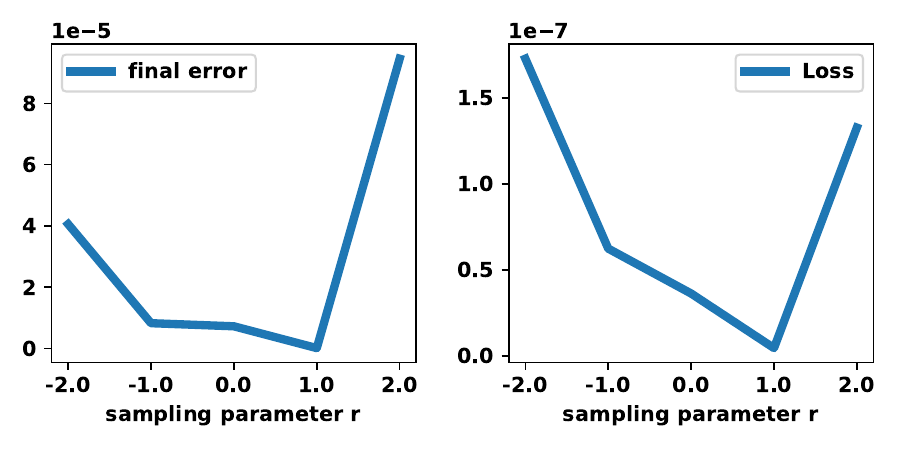}}
\caption{Sampling law influence for $\lambda=2$ (from left to right, top to bottom, plots 1 and 2),
$\lambda=-2$ (plots 3 and 4) and $\lambda=0$ (plots 5 and 6).
All sampling are done with law $\mathcal{E}^{0,T,r}$. 
Plots 1, 3 and 5 : 
 the final error as a function of $r$.
 Plots 2,4 and 6~: the loss.} \label{fig:lambda2_different_rates}
\end{figure}

We move now to the main topic which is the influence of the sampling parameter $r$ on the final error. In each case we set epoch number to $500$ (similar results are obtained for any other number of epochs) to simulate a tight computational budget and
\begin{itemize}
    \item set the $\lambda$ parameter in a list enumerating all possible regimes: negative, null or positive, here $\lambda\in \{-2,0,2\}$;
    \item compute the performance for several sampling rate $r$ parameters and look at the qualitative agreement with our theoretical results.
\end{itemize}
The numerical results are given in figure~\ref{fig:lambda2_different_rates}. 
 Each point in the plot represents a NN  
trained from scratch with the PINN loss. For consistency of the comparison each NN starts from the same Glorot initialization with the same seed.

Let start with plots $1$ and $2$ that correspond to $\lambda=2$.
It is seen that, among all possible truncated exponential laws, the one that gives minimal final error corresponds to positive value or $r$, which means that small $t$ values are given more weight. This confirms the theoretical result in proposition~\ref{prop:optimality} and  is also intuitive because here $\lambda>0$ which means exponential divergence of any perturbation.  This exponential divergence has to be corrected by an effort to solve to higher precision the early dynamics. This is also coherent with the literature, see for instance \cite{pinn_causality} that discuss the importance of causality sampling. Note that in particular this empirical results confirms that uniform time sampling, which corresponds to $r=0$ in the figure, is {\bf not optimal}.

We move now to 
plots $3$ and $4$ (figure \ref{fig:lambda2_different_rates}) that correspond to $\lambda=-2$.
 This dynamics converges to a stable equilibrium. In this case theory says that uniform sampling is not optimal and in fact giving {\bf less} weight to initial times $t$ is better because stability will erase most of the errors in this region. Note that this is {\bf at odds} with previous results from the literature that encourage oversampling for low values of $t$ irrespective of the regime. The numerical results confirm indeed that final error is minimized when sampling parameter $r$ of the truncated exponential is negative (here optimal value is $-3$). 

A special attention deserves the qualitative dependence of the loss on the sampling parameter $r$. Except for very negative $r$ values, the loss decreases with $r$ which would incorrectly suggest using large values of $r$. In fact here the loss is not informative; minimizing the loss is not the final goal, the final goal is to find the solution. 
The loss only encodes, as in reinforcement learning, the right information to find the solution. In this case, for negative values of $r$ the loss may appear larger but this happens because it works harder towards improving the outcome which is the final error. Therefore comparing two loss functionals corresponding to two different time sampling parameters $r$ will not give the expected intuitive results {\bf and will mislead the experimenter}.

Finally, the plots $5$ and $6$ (figure \ref{fig:lambda2_different_rates}) correspond to $\lambda=0$.
Here the final error seems to be low over a plateau of parameters $r$ around the value $r=0$ (optimal appears to be reached for $r=1$). So for this case the precise sampling parameter has less influence as long as it results in a somehow uniform time sampling. This is consistent with intuitive results and our theoretical results but again not always mentioned in the literature.

\subsection{Burgers' equation} \label{sec:numerical_burgers}

A test case often encountered in PINN applications is the Burgers' equation
that  describes the evolution of a one-dimensional viscous fluid flow.
This nonlinear partial differential equation reads~:
\begin{equation}
\frac{\partial u}{\partial t} + u \frac{\partial u}{\partial x} = \nu \frac{\partial^2 u}{\partial x^2}, \ u(0,x)=-\sin(\pi x), \ u(\pm 1,t)= 0 \ \forall t\in[0,1], 
\end{equation}
where $ u = u(x, t) $ is the velocity field, $ x \in [-1,1]$ is the spatial variable and $ \nu =0.01/\pi$ is the viscosity coefficient.

 Here, we consider as in \cite{RAISSI2019686}
a neural network consisting of $9$ fully connected $20$-neurons layers with 'tanh' activation. We take a space-time grid with 25 spatial points and 50 time quantiles (see previous section). We also use the trick in \eqref{eq:trick_initial_cond} to impose exact initial condition.

\begin{figure}
\centerline{
\includegraphics[width=0.55\textwidth]{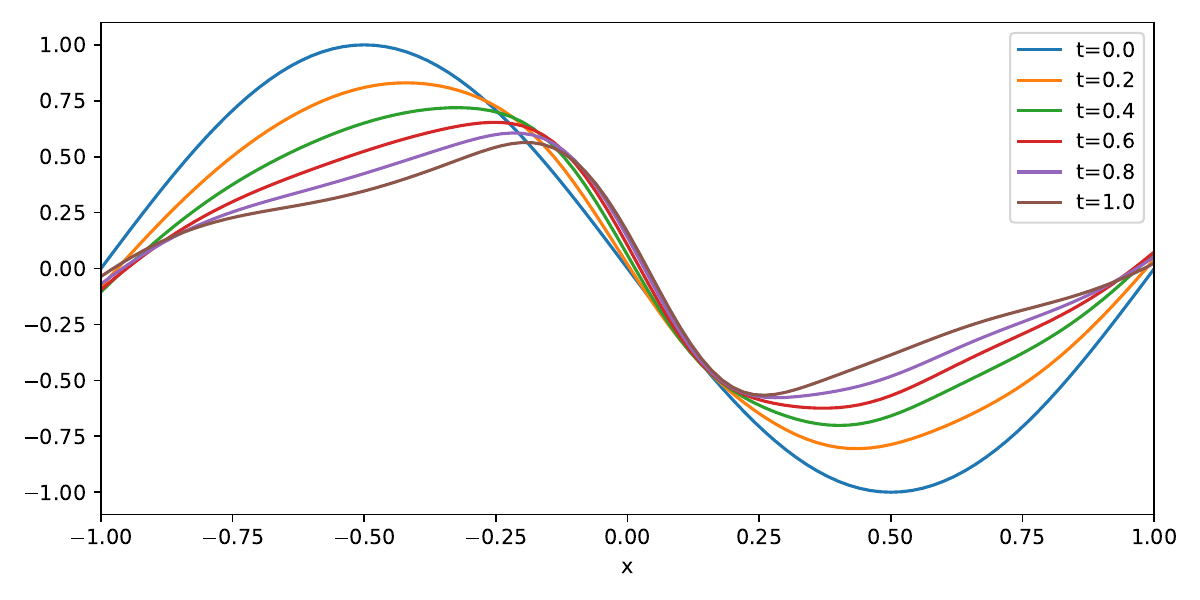}
\includegraphics[width=0.4\textwidth]{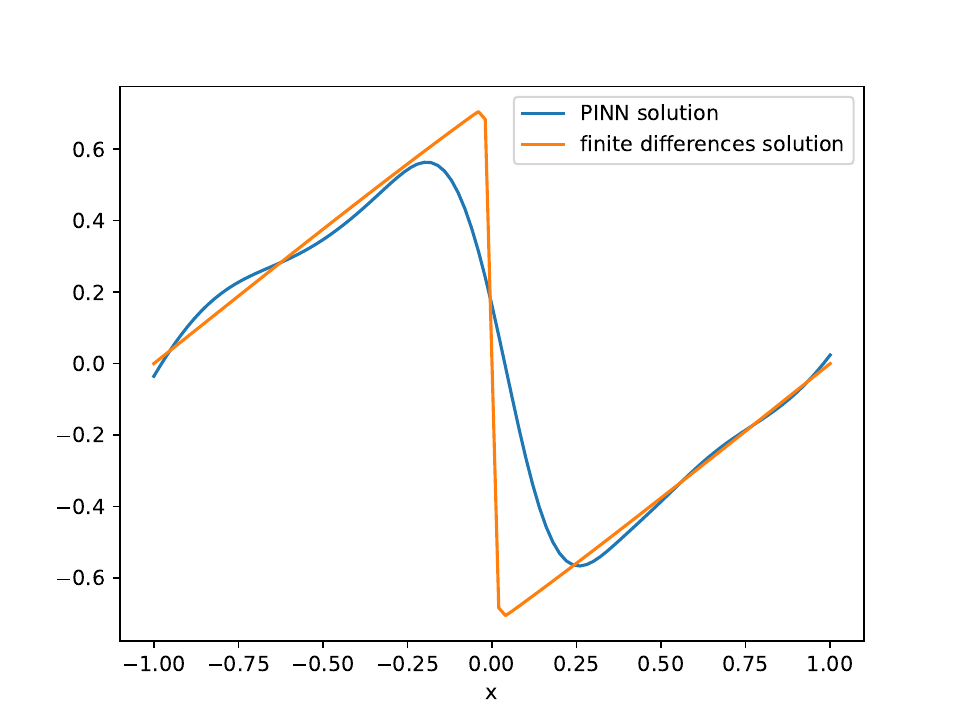}
}
\caption{Burgers' equation sampling parameter $r=0$ i.e., uniform law $\mathcal{E}^{0,T,0}$.  Left plot: the solution at different times. Right plot: the comparison with a finite difference solution considered exact.} \label{fig:burgers_lambda0}
\end{figure}

\begin{figure}
\centerline{
\includegraphics[width=0.55\textwidth]{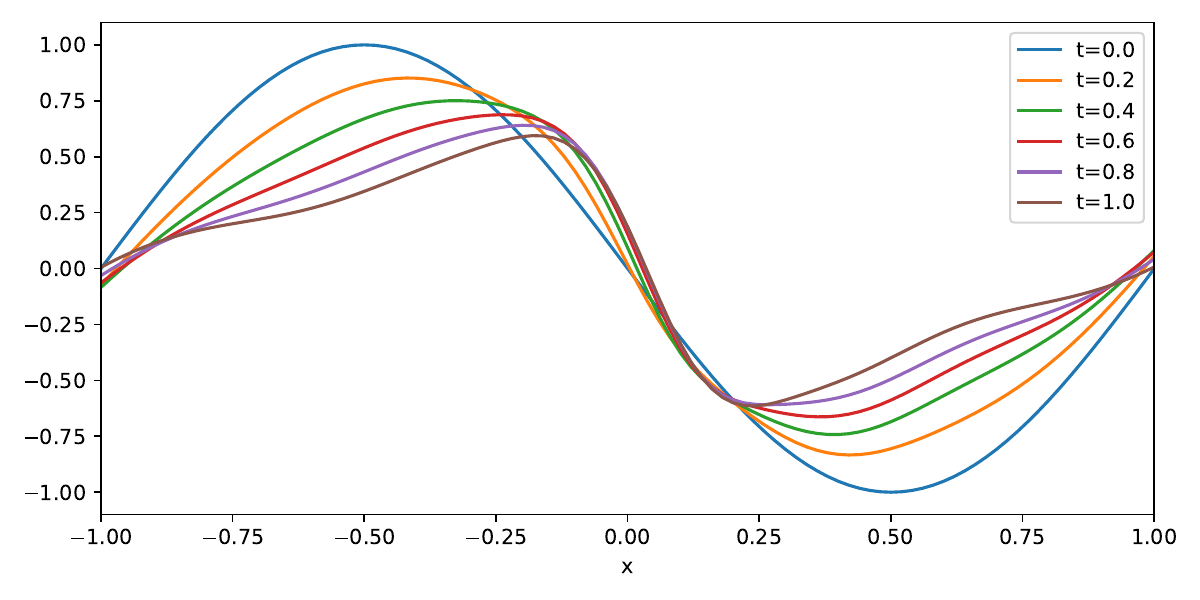}
\includegraphics[width=0.4\textwidth]{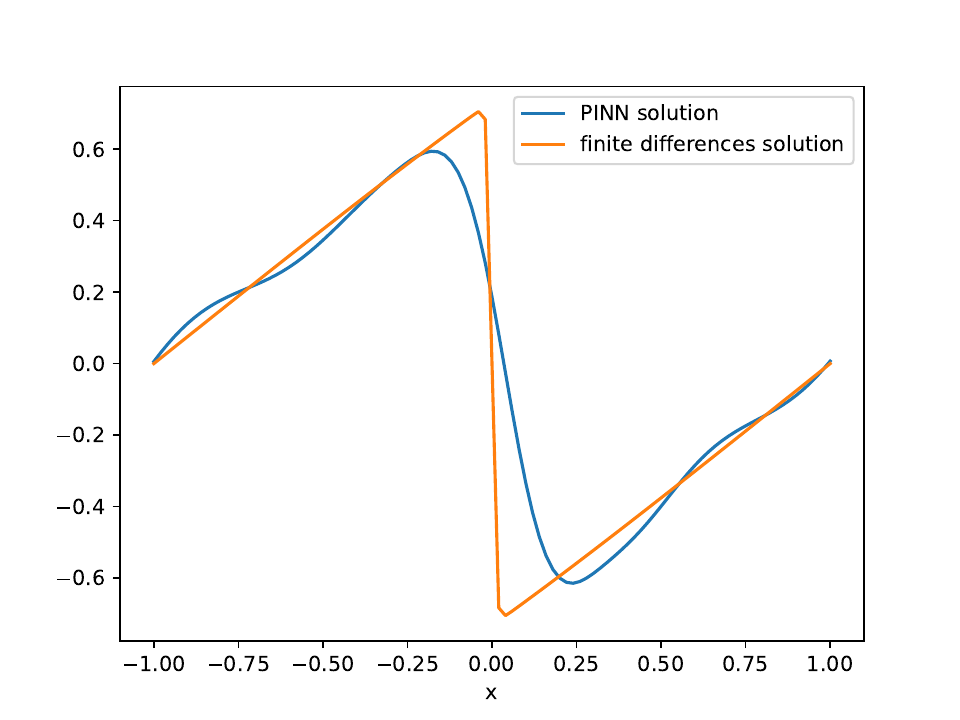}
}
\caption{Burgers' equation sampling parameter $r=1$ and law $\mathcal{E}^{0,T,r}$.  Left plot: the solution at different times. Right plot: the comparison with a finite difference solution considered exact.} \label{fig:burgers_lambda1}
\end{figure}
The results are given in figures \ref{fig:burgers_lambda0} and 
\ref{fig:burgers_lambda1}. It is seen that both laws give similar results and in practice the computation of the norm of the difference at the final time indicates that the uniform sampling is better. To explain this result we need to recall that, even if the  Burgers' equation is nonlinear and one could expect to find chaotic behavior similar
to turbulence, however, the Hopf-Cole
transformation allows to see that there  is no substantial sensitivity with respect
to initial conditions; in fact this equation is transformed to a linear parabolic equation. So the optimal sampling has no reason to overweight initial time instants and this is what we see here. 

\subsection{Lorenz system} \label{sec:numerical_lorenz}

For the final results we move to the Lorenz system, known to be chaotic, that has already been  studied in the framework of PINN \cite{pinn_causality}~:
\begin{equation}
	x'(t) = \sigma (y - x), \ \ 
	y'(t) = x (\rho - z) - y, \ \ 
	z'(t) = x y - \beta z.
\end{equation}

Here $ x, y, z $ are the state variables and $ \sigma, \rho, \beta $ are  parameters~: $ \sigma $ is the Prandtl number, $ \rho $ is the Rayleigh number, $ \beta $ is a parameter related to the aspect ratio of the system. We take as in \cite{pinn_causality} $\sigma=10$, $\rho=28$, $\beta=8/3$ and initial state $(1,1,1)$.

To be coherent with previous implementations, instead of sampling under the law $\mathcal{E}^{0,T,r}$ we take uniform sampling but use the density of $\mathcal{E}^{0,T,r}$ as weight i.e. use time weighting
proportional to $e^{-rt}$. We use the same NN as in section~\ref{sec:numerical_lambda} but with $20$ neurons per layer, $10'000$ iterations and shifting trick in \eqref{eq:trick_initial_cond}. The results are given in figure 
\ref{fig:lorenz0}. 
In this case the uniform i.e., $r=0$ weighting does not manage to obtain a reasonable solution.
 This is understood from the fact that the equation error remains large at the initial times and, due to the chaotic behavior of the system, divergence with respect to the correct trajectory will occur. The same happens for $r=-10$ which over-weights the final time instants as the expense of the initial ones. 
 On the contrary, putting more weight on the initial time instants as is done in figure~\ref{fig:lorenz0} (bottom three plots) will bring the numerical solution very close to the exact solution. 

\begin{figure}
\begin{center}
\raisebox{1.5cm}{$r=0$ \hspace{0.5cm}}
\includegraphics[width=0.85\textwidth]{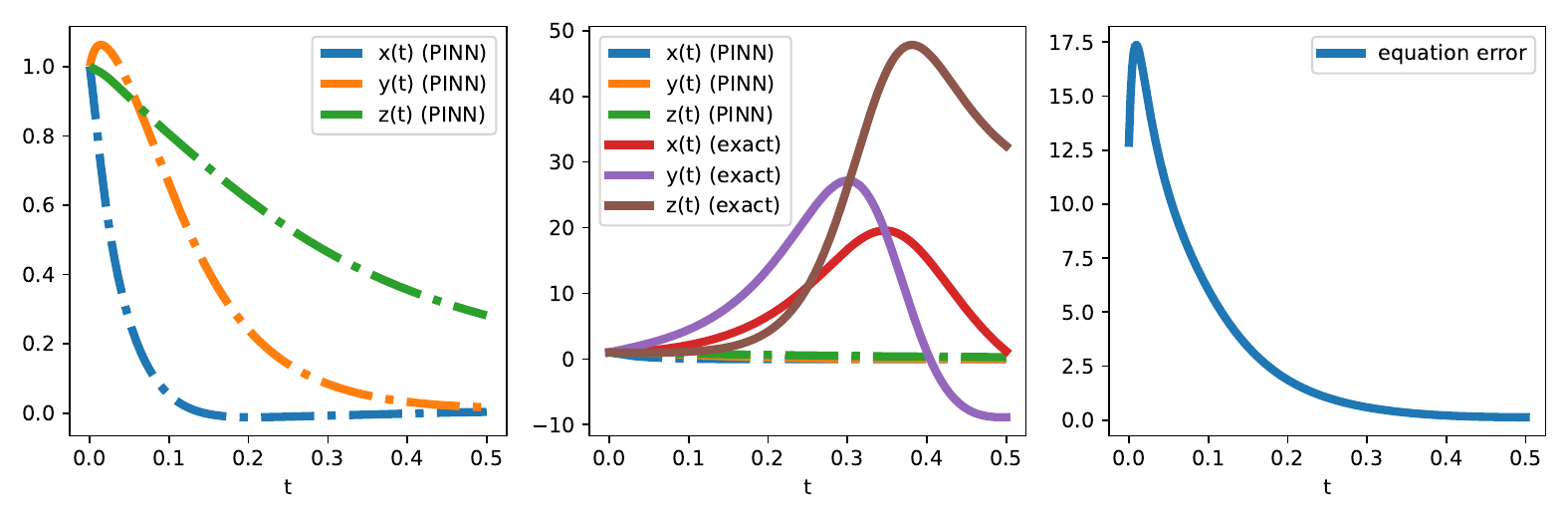}
\end{center}
\begin{center}
\raisebox{1.5cm}{$r=-10$ \hspace{0.05cm}}
\includegraphics[width=0.85\textwidth]{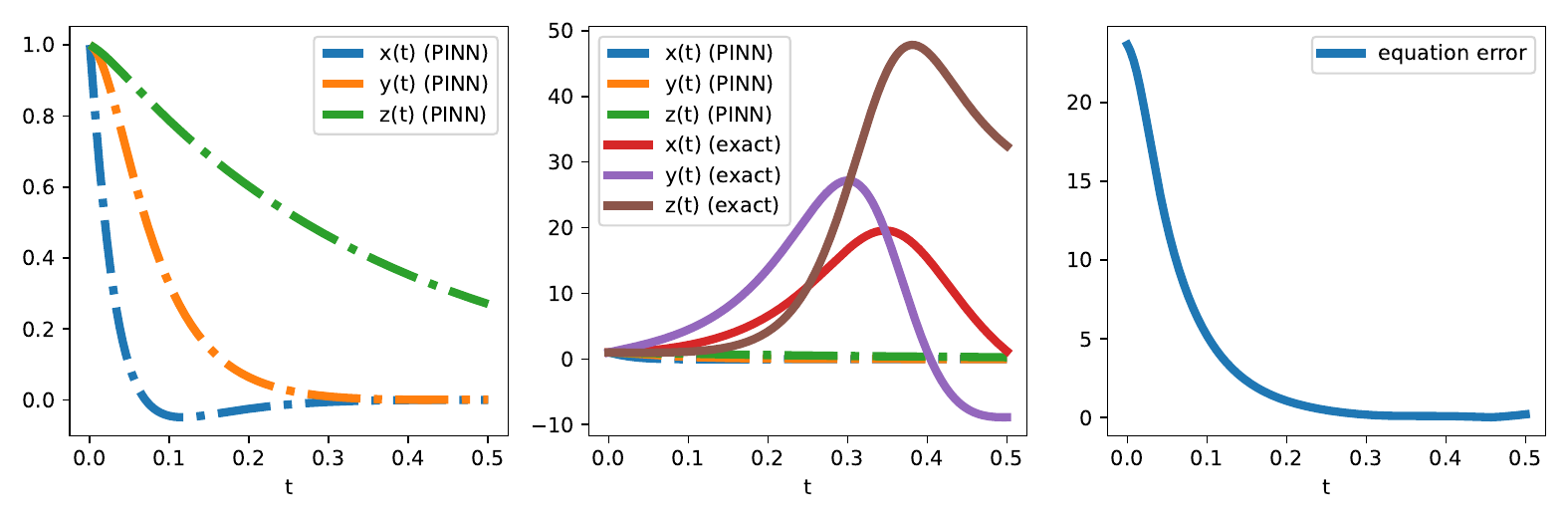}
\end{center}
\begin{center}
\raisebox{1.5cm}{$r=10$ \hspace{0.275cm}}
\includegraphics[width=0.85\textwidth]{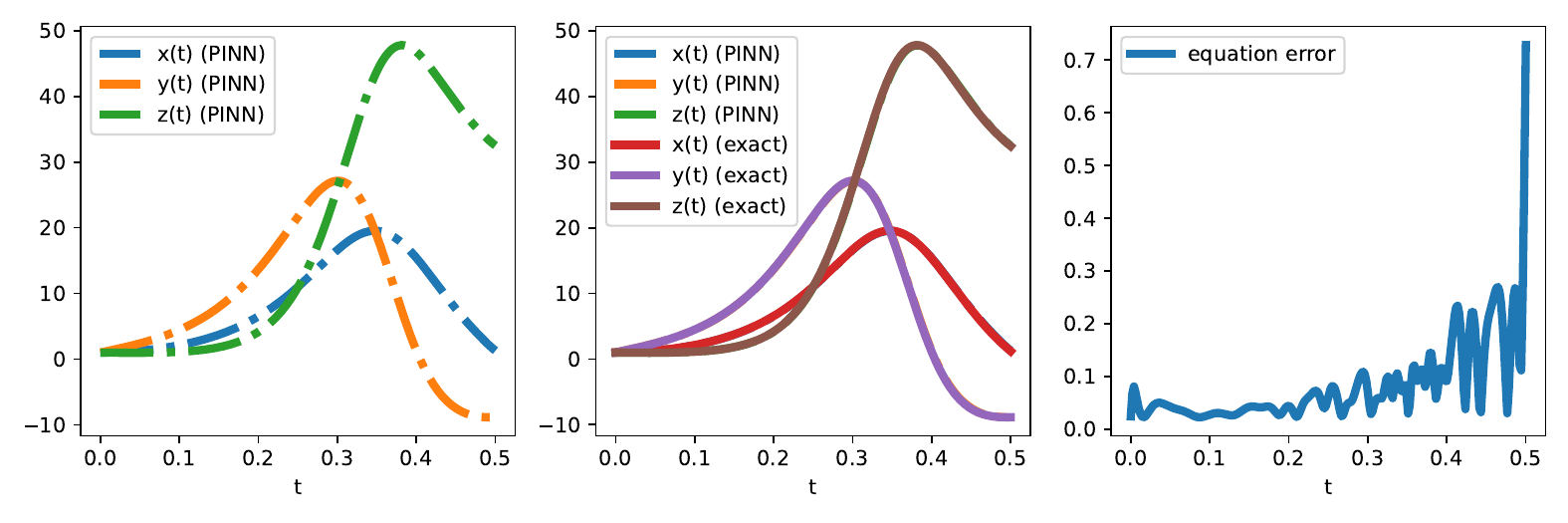}
\end{center}
\caption{Lorenz system with weight parameters $r=0$ (first row of plots), $r=-10$ (second row of plots) and $r=10$ (third row of plots). The approximation quality is not good for $r=0$ and $r=-10$ and much improved for $r=10$, in fact the numerical and exact solutions are superposed and indistinguishable graphically.} 
\label{fig:lorenz0}
\end{figure}

\section{Discussion and conclusion} \label{sec:discussion}

The goal of the PINN framework is to find the solution to a given evolution equation. This goal is transcribed through the use of a loss functional. Many loss functionals give, in the limit of infinite computational budget, the same optimal solution. But in practice the computational budget is limited and not all loss functional behave alike. We discuss here the effect of the temporal sampling on the error of the solution at the final time; to this end we prove for the first time that, under hypothesis regarding the optimization algorithm, the optimal sampling belongs to the class of truncated exponential distribution. 
We additionally characterize the optimal distribution parameter. The qualitative insight is that when the evolution is chaotic or sensitive to initial conditions early time instants should be given more weight (exponentially). On the contrary when the evolution is periodic or stably converging to an equilibrium this over-weighting is not useful any more. The theoretical results were checked numerically on several important examples and the empirical observations are coherent with them.

The principal limitation of the work is that the optimal sampling parameter is  in general unknown and has to be selected in the usual manner of hyper-parameter search. Future work will hopefully shed some light on what is the best practice to reach this optimal sampling regime.

\appendix
\section{Appendix : truncated exponential distribution}
A truncated exponential distribution is described by a triplet of parameters $(a,b,r)$, $a\le b$; the parameters $a$ and $b$  define the support of the distribution $[a,b] \subset \R$ while the rate $r$  defines the speed of decay. The distribution is by definition the only probability measure 
$\truncexp^{a,b,r}$
with support in $[a,b]$ and density proportional to $e^{-r t}$, i.e.,
\begin{equation}
\truncexp^{a,b,r}(dt) = r \frac{e^{-r t}}{e^{- r a} - e^{-r b}} \one_{[a,b]} dt.
\label{eq:def_truncated_exp}
\end{equation}
Note that it is not required that $r\ge 0$. When $r\to 0$ we obtain the uniform distribution on $[a,b]$ denoted $U(a,b)$. When $a=0, b=\infty$ we obtain the (non-truncated) exponential distribution of rate $r$.
To sample from this law, direct computations allow to show that~\footnote{Here $X \sim \mu$ means that the random variable $X$ follows the law $\mu$.}:
\begin{equation}
    \text{If } U \sim U(0,1) \text{ then }
    Y= \frac{- \ln(1-U+Ue^{-r(b-a)})}{r} \sim \truncexp^{a,b,r}.
\end{equation}

In particular the $q$-quantile of this distribution is precisely 
\begin{equation}
\frac{- \ln(1-q+qe^{-r(b-a)})}{r}. 
\label{eq:quantiles_truncated_exp}
\end{equation}
\section{Appendix : further quality metrics}

One could ask what happens when our main output is not the solution at final time $T$ but some integral over all times, i.e., instead  of $\delta u(T)$ our quality metric is~:
\begin{equation}
\int_0^T |\delta u(t)| dt = \int_0^T |\Ucal_\theta(t) - u(t)| dt.    
\label{eq:metric_0toT}
\end{equation}
This is answered in the following result.
\begin{proposition}
Under the hypothesis {\bf (H Opt)}
 the minimization problem corresponding to the loss $L_\rho$ in \eqref{eq:loss_rho} and the computational constraint
\eqref{eq:error_definition}
is guaranteed to obtain the best error level for the metric \eqref{eq:metric_0toT}
 only when 
\begin{equation}
 \rho(t) = \frac{ (e^{-\lambda t}- e^{-\lambda T})^{4/3}}{ \int_0^T (e^{-\lambda t}- e^{-\lambda T})^{4/3}}. 
 \label{eq:density_solution0toT}
\end{equation}
\label{prop:optimality_0toTmetric}
\end{proposition}
Note that although the density in 
\eqref{eq:density_solution0toT} is not exactly a truncated exponential, it will became one in the limit $T\to \infty$. 

\noindent {\bf Proof~:} 
Many parts of the proofs are the same as soon as we recognize that, under same hypothesis, the 
formula of the error metric 
$\delta u(T)$ 
given in 
\eqref{eq:integral_form_of_errorT}
can be replaced by 
\begin{eqnarray}
& \ & 
\int_0^T |\delta u (t)|dt  = 
\int_0^T \int_0^t e^{\lambda(t-s)}w(s)ds dt 
= 
\int_0^T \frac{e^{\lambda(T-s)}-1}{\lambda} w(s)ds. 
  \label{eq:integral_form_of_error0toT}
\end{eqnarray}
The $w$ that minimizes \eqref{eq:integral_form_of_error0toT} under resources constraint \eqref{eq:error_definition} is found as before to be proportional to 
$(e^{\lambda(T-s)}-1)^{1/3}$. The rest follows as before.

\bibliographystyle{splncs04}
\bibliography{optimal_time_sampling_pinn}
\end{document}